\documentclass[letterpaper, 10 pt, conference]{ieeeconf}  

\IEEEoverridecommandlockouts                              

\usepackage{amsmath} 
\usepackage{bbding}
\usepackage{pifont}
\usepackage{wasysym}
\usepackage{amssymb}

\usepackage{graphicx}
\usepackage{bbding}
\usepackage{pifont}
\usepackage{wasysym}
\usepackage{listings}
\usepackage{zebra-goodies}

\title{\LARGE \bf
SaaF: Scene-Specific Ambiguity-Aware 3D Language Fields \\towards Interactive Real-World Object Retrieval
}

\author{Yuga Yano$^{1, 2}$, Daiju Kanaoka$^{2}$, Hakaru Tamukoh$^{2, 3}$ and Yasutomo Kawanishi$^{2}$
\thanks{This work was supported by JST SPRING, Japan Grant Number JPMJSP2154. Parts of this research were supported by MEXT, Grant-in-Aid for Scientific Research (24H00733).}
\thanks{$^{1}$ Kyushu Institute of Technology, Fukuoka, Japan.}
\thanks{$^{2}$ RIKEN, Kyoto, Japan.}
\thanks{$^{3}$ Research Center for Neuromorphic AI Hardware, Fukuoka Japan.}
\thanks{For correspondence:~{\tt\small\ yano.yuuga158@mail.kyutech.jp}}
}
\usepackage{amsmath}
\usepackage{booktabs}
\usepackage{multirow}
\usepackage{cleveref}
\usepackage{url}
\crefname{figure}{Fig.}{Figs.}
\Crefname{figure}{Fig.}{Figs.}

\begin{document}

\maketitle
\thispagestyle{empty}
\pagestyle{empty}

\begin{abstract}
We propose Scene-specific Ambiguity-aware 3D Language Fields (SaaF), a novel Gaussian Splatting-based 3D language field designed for interactive object retrieval in a given real-world scene.
Interactive object retrieval using natural language is a crucial capability for service robots operating in complex real-world environments. 
While recent 3D language field methods for object retrieval establish associations between rendered pixels and autoencoder-compressed CLIP features, they suffer from two limitations: (1) reduced discriminability among similar objects due to feature compression, and (2) poor handling of ambiguous queries, often resulting in unstable or incorrect retrieval.
To address these limitations, SaaF introduces a metric learning strategy to construct a unified feature space that is both instance-discriminative and ambiguity-aware.
(i) To enhance instance-level visual discrimination, SaaF employs metric learning that pulls image features from multiple viewpoints of the same object closer together in the feature space.
(ii) To establish ambiguity awareness, the model jointly trains on multiple text labels generated by the proposed method from each tracked object image sequence, including ambiguous descriptions, to learn the semantic relationships between ambiguous and specific features in a target scene.
This feature space enables fine-grained visual understanding while allowing the system to estimate query ambiguity and interactively request clarification when needed.
Experimental results demonstrate that SaaF not only improves retrieval accuracy over previous methods but also robustly detects and handles ambiguity in the user text queries under open-vocabulary settings.
\end{abstract}


\section{Introduction}
Robots are increasingly expected to follow natural language instructions and to plan actions to complete tasks~\cite{zitkovich2023rt,yano2024unified}.
In real-world scenarios, user instructions are often ambiguous.
KnowNo~\cite{ren2023robots}, an interactive task planning system built on PaLM-E SayCan~\cite{ahn2022can}, addresses this challenge by detecting ambiguous instructions and proactively querying the user for clarification.
The ability to assess whether a given instruction is sufficiently clear, and to decide whether further interaction is required, is essential for the practical use of robots.
However, KnowNo assumes that object locations and names are already known, and it does not address detecting objects from user queries (open-vocabulary object retrieval).
Therefore, open-vocabulary object retrieval~\cite{azuma_2022_CVPR,sim_vqa_2022} is an essential capability for service robots operating in complex environments.
In this study, we focus on detecting and resolving query ambiguity in interactive scenarios.

\begin{figure}[t]
    \centering    \includegraphics[width=1.\linewidth]{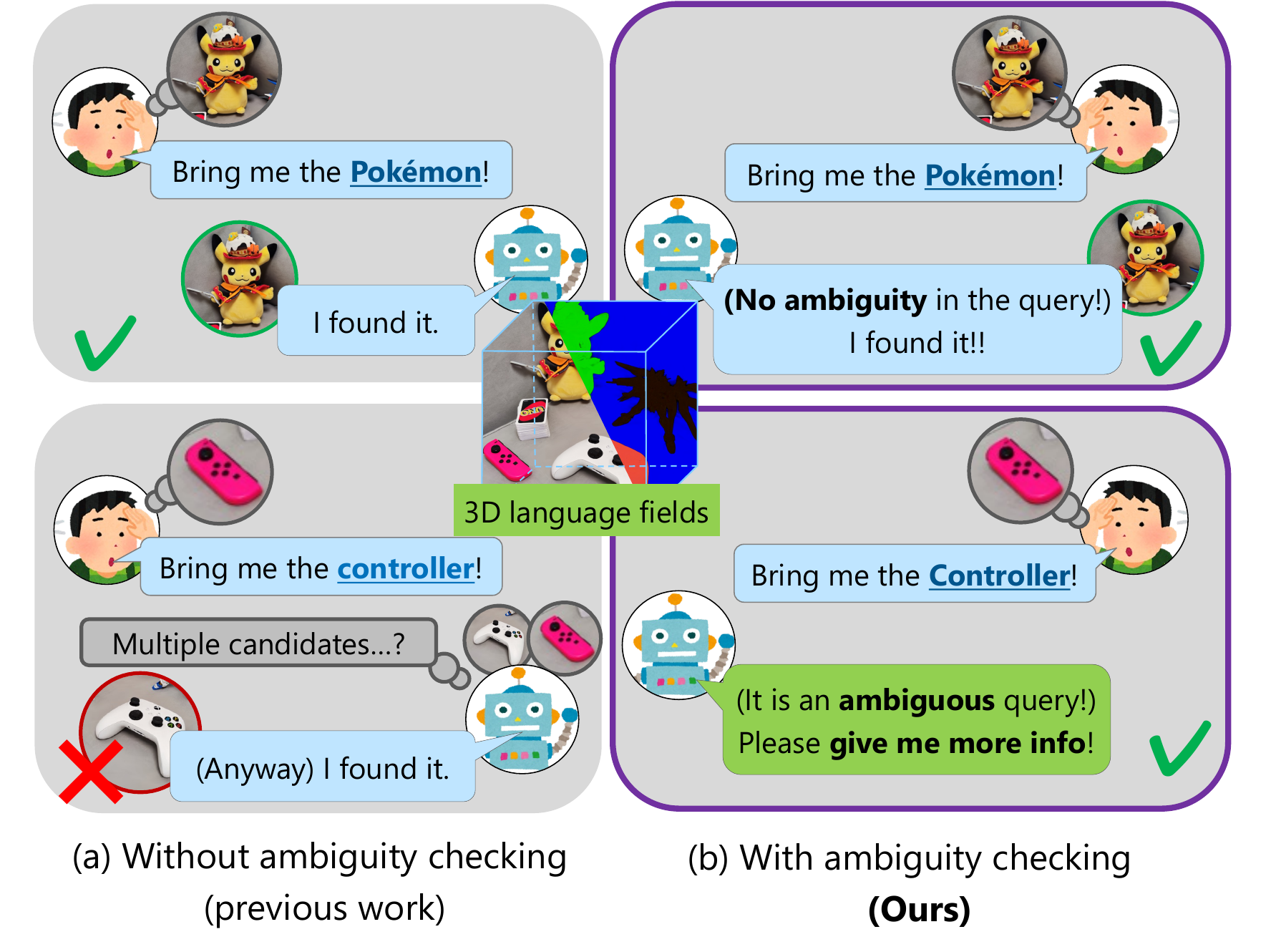}
    \caption{\textbf{Target scenario}: Conventional 3D language fields only focus on retrieving the most relevant object within a scene and do not account for scenarios where multiple possible candidates exist. We propose a novel 3D language field that simultaneously checks query ambiguity and retrieves an object. Our method quantitatively evaluates query ambiguity before retrieving objects and requests additional user instruction when the query is ambiguous. }
    \label{fig:target_scnario}
\end{figure}

For object retrieval based on a natural language query, it is necessary to associate the objects features in the environment with the semantic features of the user query.
In robotics, 3D reconstruction has recently attracted attention as a method for spatial perception, since it enables the robot to represent a physical environment in a unified 3D space.
Based on this progress, several methods have been proposed to embed language features into reconstructed 3D scenes using models such as CLIP (Contrastive Language-Image Pre-training)~\cite{clip_pmlr-v139-radford21a}.
A reconstructed 3D scene in which each spatial element is associated with language features is referred to as a 3D language field.
Recently, 3D language fields based on Gaussian Splatting~\cite{kerbl3Dgaussians}, such as LangSplat~\cite{Qin_2024_CVPR}, have been proposed~\cite{shi2024language,Qin_2024_CVPR}.
LangSplat achieves open-vocabulary object retrieval by embedding autoencoder-compressed CLIP features into each Gaussian that represents the reconstructed 3D scene.
The method extracts object masks in all the input images using Segment Anything (SAM)~\cite{kirillov2023segment} and computes CLIP features from each mask.
An autoencoder is then trained on the extracted CLIP features to obtain a low-dimensional compressed feature space.
The compressed features reduce the computational cost of feature retrieval and help suppress noisy variations.
However, since the autoencoder is trained solely to minimize reconstruction error, it does not explicitly preserve differences in CLIP features of different object instances.
As a result, features of visually similar but distinct object instances tend to become less separable in the compressed space.
This limitation can lead to unstable or inconsistent retrieval results in LangSplat, particularly when user instructions are ambiguous, as shown in Fig. 1(a).

To address these issues, we propose a novel 3D language fields model that enables ambiguity estimation and is well-suited for scene-specific object retrieval, allowing discrimination between objects within a targeted real-world scene.
\Cref{fig:target_scnario}(b) shows an overview of our target scenario, interactive real-world object retrieval. 
First, our method reconstructs the 3D space of the target environment using Gaussian Splatting.
Then, scene-specific features are embedded into the reconstructed 3D Gaussians.
To obtain these features, we introduce text-label generation with multi-level ambiguity for each object instance based on object tracking and a vision-language model (VLM), along with a metric-learning-based autoencoder that compresses CLIP features while explicitly preserving differences between object instances and between ambiguous and specific queries.
This training strategy enables a scene-specific and ambiguity-aware feature space.
After constructing the 3D language field, the user inputs a text query describing an object to retrieve.
A text feature is extracted from the query and evaluated in the learned feature space to determine whether the query is ambiguous within the scene.
If the query is ambiguous, the system provides feedback and waits for a more specific instruction.
Once the system determines through the interaction that the query is sufficiently specific for retrieval in the scene, it identifies and outputs the region that matches the textual features.

To demonstrate the effectiveness of SaaF, we conducted experiments on several datasets for OVS.
Experimental results demonstrate that SaaF outperforms state-of-the-art methods for 3D language fields in terms of object retrieval accuracy.
Furthermore, we demonstrated that SaaF can detect ambiguous queries by evaluating the alignment in the feature space.

Our main contributions in this study are as follows:
\begin{itemize}
    \item We present SaaF, a 3D language field that embeds scene-specific features into Gaussian Splatting, enabling interactive object retrieval.
    \item We introduce a learning strategy that constructs an ambiguity-aware feature space by leveraging text labels with varying levels of ambiguity generated by object tracking and a Vision-Language Model (VLM).
    \item We demonstrate through experiments on multiple datasets that the proposed method acquires a feature space effective for both precise object retrieval and ambiguity evaluation.
\end{itemize}

\section{Related Works}
Since 3D language fields rely on 3D representations, this section first summarizes recent advances in 3D scene representations, followed by a review of 3D language fields.
Finally, we introduce visual foundation models used for extracting image and text embeddings.

\subsection{3D Representations}
Neural Radiance Fields (NeRF)~\cite{nerf} have emerged as a groundbreaking approach for synthesizing images from novel views of a 3D scene from multiple images.
By training a multi-layer perceptron (MLP) to map 3D coordinates and viewing directions to radiance and density values, NeRF implicitly represents a scene as a continuous volumetric field.
However, as an implicit representation, NeRF makes it difficult to decompose a scene into individual components or objects, limiting its applicability to downstream tasks such as editing or semantic understanding.
Moreover, since NeRF relies on random sampling, it leads to slow rendering times and makes real-time applications challenging.

To address these limitations, Gaussian Splatting~\cite{kerbl3Dgaussians} has been introduced. 
Gaussian splatting represents a 3D scene using a set of colored 3D Gaussians, which are projected onto image planes and rendered via alpha blending.
Its explicit structure allows individual Gaussians to be independently manipulated, making the representation well-suited for downstream tasks such as segmentation, editing, and semantic reasoning. 
Furthermore, Gaussian Splatting offers significant advantages in terms of training speed and real-time rendering capabilities.
Notably, methods such as Gaussian Grouping~\cite{ye2024gaussian} extend this framework by assigning identity encodings to Gaussians, enabling object-level grouping and fine-grained scene understanding.

In summary, while NeRF excels at high-quality reconstruction, it lacks efficiency and interpretability.
Gaussian splatting addresses these issues, offering a more practical and flexible solution for 3D scene reconstruction and manipulation.

\begin{figure*}[t]
    \centering
    \includegraphics[width=\linewidth]{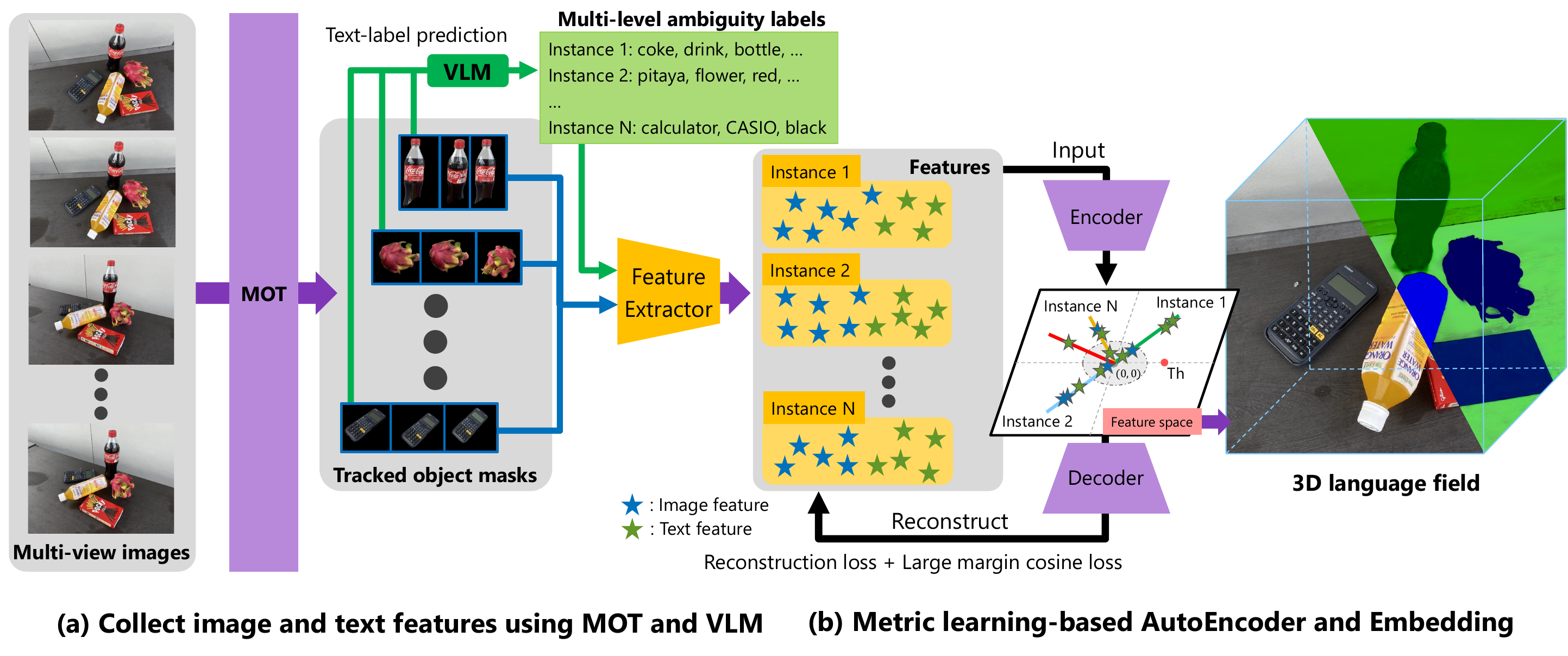}
    \caption{\textbf{Proposal overview.} The proposed method, SaaF, consists of the following two components. One is the generation of tracked object masks using MOT and text labels with multiple ambiguities utilizing a Vision-language model. The other is a metric learning-based autoencoder that learns a feature space for scene-specific object retrieval and embeds these features into 3D fields. The autoencoder is trained to both discriminate between object instances and detect ambiguous queries based on the feature space.}
    \label{fig:proposal_overview}
\end{figure*}

\subsection{3D Language Fields}
LERF~\cite{lerf2023} is a representative NeRF-based method that embeds language features into the 3D space, pioneering the concept of 3D language fields using embedded CLIP features.
It attracted attention for enabling open-vocabulary queries within reconstructed scenes, and subsequent research has extended this framework by embedding features such as DINO~\cite{caron2021emergingDINO} and applying it to robotic tasks.
However, NeRF-based 3D language fields are limited in real-world robotics due to their high computational cost for training and rendering processes.

To address these limitations, Gaussian splatting-based 3D language fields~\cite{wu2024opengaussian,shi2024language,Qin_2024_CVPR} have been proposed.
These methods embed image-text features into Gaussians corresponding to object regions segmented by SAM, enabling object-level language-based retrieval.
Approaches such as Segment Any GAussains (SAGA)~\cite{saga2023}, Click-gaussian~\cite{choi2024click}, and Gaussian Grouping~\cite{ye2024gaussian} construct 3D language fields by embedding the unified features into Gaussians corresponding to object regions segmented by SAM.
In these approaches, Gaussians representing the same object share identical features within the reconstructed space, allowing users to extract an entire 3D object region by selecting a certain Gaussian.

However, these methods do not embed CLIP features directly into the 3D field.
Instead, to perform open-vocabulary object detection, they render segmented object regions from a specific viewpoint and extract CLIP features from the rendered image.

LangSplat~\cite{Qin_2024_CVPR} is a method to embed low-dimensional CLIP features~\cite{clip_pmlr-v139-radford21a}, compressed via an autoencoder, into each point.
This compression surpasses feature variance and reduces the computational cost of feature embedding.
However, the compressed features tend to make visually similar objects more difficult to distinguish, as the autoencoder reduces small differences in the original CLIP space.
Furthermore, LangSplat can fail to detect the correct object or may detect multiple objects when the language query is ambiguous, such as referring to multiple objects.
In tasks such as object search or grasping in robotics, it is crucial to accurately classify visually similar objects and to assess the ambiguity of the given query.

\subsection{Visual Foundation Models}
CLIP~\cite{clip_pmlr-v139-radford21a} is a visual foundation model that embeds both images and text into a shared feature space.
Many applications leveraging CLIP features have been proposed, including zero-shot image recognition~\cite{kirillov2023segment,ravi2024sam2} and open-vocabulary object detection (OVD)~\cite{liu2024grounding,li2022languagedriven}, in which objects are detected using textual queries.

In 2023, Meta introduced the Segment Anything Model (SAM)~\cite{kirillov2023segment}, which improves zero-shot object detection performance by leveraging CLIP features.
SAM2~\cite{ravi2024sam2}, released in 2024, further enhances recognition accuracy and supports multi-object tracking (MOT) by introducing a memory bank.
In this study, we use SAM2 with MOT to obtain segmented masks for each object and consistent instance labels across frames, which are used to train our metric learning- based autoencoder.

Additionally, vision language models (VLMs) such as InstructBLIP~\cite{dai2023instructblip}, Llava~\cite{liu2023visual}, and GPT-4v~\cite{gpt4v}, have attracted much attention for their ability to describe images in natural language, detect objects in images, and generate images from textual input.
In our work, we use VLMs to generate textual labels of varying levels of ambiguity, which include supercategories for objects in images.
These labels serve as auxiliary semantic information during training.


\section{Scene-Specific Ambiguity-aware 3D Language Field}
Previous 3D language fields, such as LangSplat, face challenges: feature compression using a reconstruction-based autoencoder makes it difficult to classify visually similar objects, and ambiguous queries make it hard to identify the target object.
To address these issues, we propose a model and its training strategy that can both determine between object instances and assess the ambiguity of queries.
\Cref{fig:proposal_overview} shows the proposal overview of the proposed 3D language field.

To realize scene-specific 3D language fields, SaaF consists of image and text feature collection shown in \cref{fig:proposal_overview}(a) and metric learning-based autoencoder training shown in \cref{fig:proposal_overview}(b).
\Cref{sec:feature_collection} describes a feature collection with tracked object masks using MOT, and generates various text labels through VLM.
\Cref{sec:learning_and_embedding} presents the training strategy of the autoencoder and embedding compressed features to a 3D language field.
In addition, \cref{sec:interactive_object_retrieval} presents methods for assessing ambiguity and detecting the target object in an interactive scenario.

\subsection{Image and Text Feature Collection}
\label{sec:feature_collection}

\subsubsection{\textbf{Object mask tracking}}
For training an autoencoder based on metric learning, the proposed method requires sets of object regions for object instances in the scene.

SaaF collects tracked object masks with consistent instance IDs by a MOT system, as shown in \cref{fig:proposal_overview}(a).
The proposed method segments on the first frame into object regions and performs multi-object tracking based on the detected masks for subsequent frames.
The proposed method is also applied to the subsequent frames to detect new objects that are not tracked.
The IoU of the detected and tracked object regions, and adds the detected masks that have with less than a certain overlap with existing tracks as new objects.
SaaF repeats this tracking process for all frames to create tracked object masks with consistent instance IDs.
In this paper, we used SAM2 as the MOT system and CLIP as a feature extractor.

\subsubsection{\textbf{Multiple Ambiguity Text Label Generation}}
\label{sec:label_generation}
For training an autoencoder with text features of varying levels of ambiguity, the proposed method requires sets of text labels per object instance in the scene.

\Cref{fig:proposal_overview}(a) also shows how the proposed method generates text labels by utilizing tracked mask images.
The proposed method combines multiple frames of tracked object regions into an image and inputs the combined image to VLM to generate about 100 different text labels for each object instance.
The masked object images are sorted in descending order based on mask area size, and several top-ranked frames are randomly selected.
The randomness is introduced to avoid similar masks and to prevent generated text labels from being biased, which could arise if only the masks from the top few adjacent frames are used.
The text labels generated by VLM contain words with varying levels of ambiguity, such as the object attributes (e.g., color and shape), common and unique names, and supercategories.
In this study, we use GPT-4o as a VLM to generate text labels.
The system prompts given to GPT-4o to make text labels are as follows:
\begin{lstlisting}[basicstyle=\ttfamily\footnotesize, breaklines=true, frame=single]
System prompt for generate text labels: 
Please output various names that accurately expresses the characteristics of this object. If proper nouns such as product names are known, they should also be output as label names. Label names should include the shape, color, and other distinguishing characteristics. If you know this character, create some labels include the unique character name. When manufacturer's or brand's name or logo is written on the product, you have to include that in some labels. Do not focus on the blacked-out silhouettes, but only on the colored areas to generate labels.

System prompt for generate ambiguous labels:
Please make about 100 labels for each object. Create some labels that focuses on the super category of the object.
\end{lstlisting}
In order to generate multiple ambiguous text labels, we add a focus on supercategories and an increase in the number of generated labels to VLM's system prompts.

\subsection{Metric Learning and Feature Embedding}
\label{sec:learning_and_embedding}

\subsubsection{\textbf{Learning scene-specific feature space}}
We propose a metric learning-based autoencoder to learn a feature space that enables scene-specific object retrieval and ambiguity determination, by using the collected image and text features explained in \cref{sec:feature_collection}.

\begin{figure}[t]
    \centering
    \includegraphics[width=\linewidth]{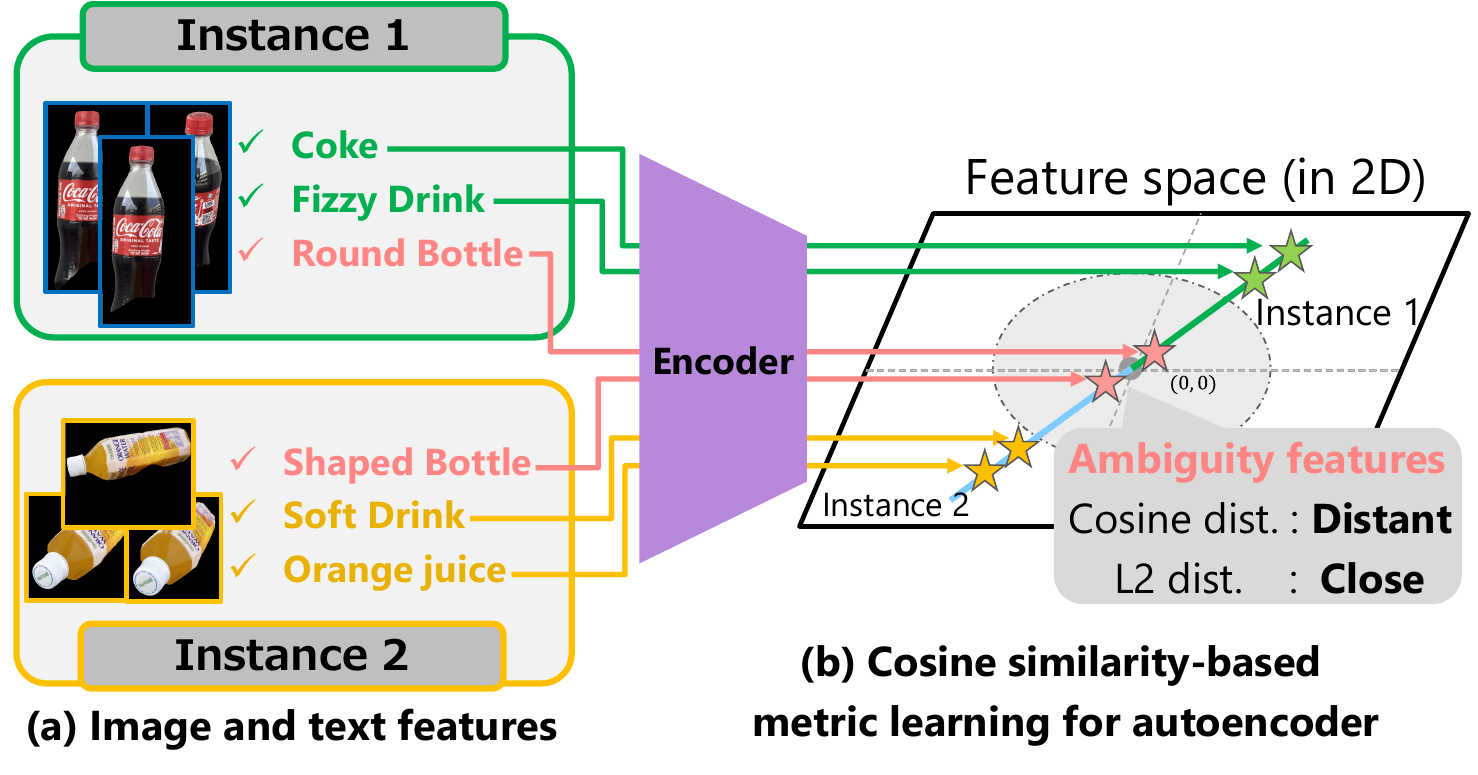}
    \caption{\textbf{Feature space of the SaaF.} In the proposed metric learning, unique features such as \texttt{Coke} and \texttt{Orange juice} are mapped to distant positions in the feature space. On the other hand, ambiguous features such as \texttt{Round bottle} and \texttt{Shaped bottle} that span both instances are mapped close to the origin to satisfy both distance constraints and reconstruction accuracy.}
    \label{fig:autoencoder_feature_space}
\end{figure}

\Cref{fig:proposal_overview}(b) shows an overview of the proposed metric learning-based autoencoder.
We propose a loss function consisting of cosine similarity-based loss ($\mathcal{L}_{\text{lmc}}$) and reconstruction loss ($\mathcal{L}_\text{reconst}$) to learn a feature space as shown in \cref{fig:autoencoder_feature_space}.
\begin{equation}
\mathcal{L} = \mathcal{L}_{\text{lmc}} + \lambda\mathcal{L}_{\text{reconst}},
\end{equation}
where $\lambda$ denotes a balance parameter.
Using the cosine similarity-based loss $\mathcal{L}_{\text{lmc}}$ aims to increase cosine similarity between features of the same instance and to reduce the similarity between features of different instances.
Given a set of images $\mathcal{I}_i$ of the $i$-th tracked object instance, a set of CLIP features $\mathcal{F}_i$ is extracted.
Subsequently, these features are embedded in the trained feature space by using the autoencoder, and set of embedded features ${\mathcal{E}}_i$ is obtained.
$\widehat{\mathcal{F}}_i$ is also obtained from the compressed features ${\mathcal{E}}_i$ via decoder.
Based on the given sets of image features ${\mathcal{E}}_i$ and ${\mathcal{E}}_j$, the $\mathcal{L}_{\text{lmc}}$ loss are computed as
\begin{align}
    \mathcal{L}_{\text{lmc}} &=  \mathcal{L}_{\text{pos}} + \mathcal{L}_{\text{neg}},\\
    \mathcal{L}_{\text{pos}} &= 1 - a,\\
    \mathcal{L}_{\text{neg}} &= \max(M - b,\, 0),
    \label{eq:lmc_fin}
\end{align}
where $M$ is a certain margin.
$a$ and $b$ are distances of features between the same and different object instances, respectively, defined as
\begin{align}
    \label{eq:lmc_loss}
    a &= s_{\text{cos}}(\mathbf{e}_x, \mathbf{e}_y), \quad \mathbf{e}_x, \mathbf{e}_y \in \mathcal{E}_i, \\
    b &= s_{\text{cos}}(\mathbf{e}_x, \mathbf{e}_y), \quad \mathbf{e}_x \in \mathcal{E}_i,\, \mathbf{e}_y \in \mathcal{E}_j,\ i \neq j,
\end{align}
where $s_{\text{cos}}$ is a cosine similarity function.

$\mathcal{L}_{\text{reconst}}$ is calculated as the sum of the L2 and cosine distances between the reconstructed feature $\mathbf{f}_{\text{reconst}}\in\widehat{\mathcal{F}}_i$ by the autoencoder and its original CLIP feature $\mathbf{f}_{\text{orig}}\in\mathcal{F}_i$, defined as
\begin{equation}
    \label{eq:reconstruct_loss}
    \mathcal{L}_{\text{reconst}} = d_{\text{cos}}(\mathbf{f}_{\text{orig}}, \mathbf{f}_{\text{reconst}}) + d_{\text{l2}}(\mathbf{f}_{\text{orig}}, \mathbf{f}_{\text{reconst}}),
\end{equation}
where $d_{\text{cos}}$ and $d_{\text{l2}}$ are cosine and L2 distance functions, respectively.

\Cref{fig:autoencoder_feature_space} illustrates an example of the metric learning in a feature space.
Ambiguous features that overlap multiple objects, such as a round bottle and a shaped bottle, are located near the origin as a result of simultaneously satisfying both $\mathcal{L}_\text{lmc}$ and $\mathcal{L}_\text{reconst}$ constraints.
The $\mathcal{L}_{\text{lmc}}$ enforces high cosine similarity only among compressed features that represent the $i$-th object ($\mathcal{E}_i$), and low cosine similarity between different object instances.
Therefore, compressed features of the same object are aligned in a similar direction in the feature space.
In addition, to facilitate reconstruction of the AutoEncoder, similar features should be located close together in Euclidean distance in the feature space.
Therefore, ambiguous features that overlap multiple instances are learned to reduce their Euclidean distances while maintaining separable vector directions.
As a result of this learning, ambiguous features are concentrated near the origin.
In contrast, unique features are required to maintain Euclidean distance from ambiguous features near the origin, which increases the norm of the vectors.
From the above process, the feature space that determines ambiguity via vector norm is obtained.

\subsubsection{\textbf{Feature Embedding for Gaussian Splatting}}
The proposed method constructs 3D language fields by embedding compressed features into each Gaussian.
LangSplat embeds compressed features obtained from the masks in each frame into Gaussians; however, this feature embedding becomes unstable due to variations in features across frames, caused by occlusion and changes in viewpoints.
To address this, our method selects the frame with the largest mask area among those tracked by SAM2 as the best frame and uses only the features extracted from this best frame for stable embedding.

\subsection{Interactive Object Retrieval}
\label{sec:interactive_object_retrieval}
This section describes how SaaF assess ambiguity of user-provided queries and detects target objects.
Text features from user-provided queries are compressed by the trained encoder to obtain $\mathbf{q}_{\text{comp}}$.
The L2 norm of the compressed text feature $\mathbf{q}_{\text{comp}}$ is computed; if it falls below a certain threshold $\mu$, the query is classified as ambiguous, as
\begin{equation}
    \begin{cases}
    \text{ambiguous query} & \text{if } \left\| \mathbf{q}_{\text{comp}} \right\|_2 < \mu, \\
    \text{unambiguous query} & \text{otherwise}.
    \end{cases}
    \label{eq:query_type}
\end{equation}

If the norm exceeds the threshold $\mu$, the cosine similarity is calculated between the compressed query text feature $\mathbf{q}_{\text{comp}}$ and the set of embedded features $\mathcal{E}_p$. Here, each feature $\mathbf{e}_p \in \mathcal{E}_p$ corresponds to a pixel $p$ in the image rendered from the 3D language field. 
Finally, the queried object region is output as the set of pixels whose similarity score exceeds a certain threshold $\tau$:
\begin{equation}
\left\{ p \;\middle|\; s_{\text{cos}}\left(\mathbf{q}_{\text{comp}}, \mathbf{e}_p\right) > \tau,\; \mathbf{e}_p \in \mathcal{E}_p \right\}.
\label{eq:calc_similarities}
\end{equation}


\section{Experiments}

In this section, we confirm whether the proposed metric learning-based autoencoder constructs a feature space capable of discriminating objects and detecting ambiguous queries in the feature space.
We used two datasets for the experiments: the 3D-OVS~\cite{liu2023weakly} dataset and the \textit{Figurines} scene from the LERF dataset.
These datasets were chosen because they contain multiple objects that share the same supercategory, making them suitable for ambiguity evaluation.

All training and testing were conducted on a single NVIDIA RTX A6000 Ada GPU.
This includes SAM2 tracking, training the autoencoder, and Gaussian Splatting.
The parameters for metric learning are as follows:
The Encoder compresses the 512-dimensional features obtained from CLIP into 256, 128, 64, 32, and 3 dimensions using fully connected layers in a stepwise process. After each compression step, batch normalization and ReLU activation are applied. The Decoder reconstructs the compressed features into 16, 32, 64, 128, 256, 256, and 512 dimensions sequentially.
The model at the epoch with the lowest loss is adopted after training 500 epochs.
The model used Adam for optimization, the learning rate is 0.007.
We set the batch size to 1024.
PairMarginMiner was used for sampling pairs in the metric learning loss calculation.

We compared our proposed method, SaaF, against Lang-SAM (which performs OVS on single RGB images), LERF (NeRF-based), and LangSplat (Gaussian Splatting-based).
These baselines were chosen as they represent the highest performance within their frameworks.
In all cases, a featured image was rendered from a camera viewpoint, and object retrieval was performed by comparing the similarity between the CLIP features of the text query and those of the image pixels.
LangSplat reconstructs pixel-level features using a decoder before computing similarity.
In contrast, the proposed method, SaaF,  compresses CLIP features using a scene-specific encoder and directly computes the cosine similarity between the compressed features.


\subsection{IoU Accuracy}
We first evaluated segmentation accuracy in simpler scenes using the 3D-OVS dataset.
This dataset consists of approximately 30 RGB images per scene, capturing around six objects and the background.
Original image resolution is $4,032\times 3,024$ pixels, but we resized the images to $1,440\times 1,080$ pixels for our experiments, following the experimental settings used in LangSplat.

\begin{figure*}
    \includegraphics[width=1.0\linewidth]{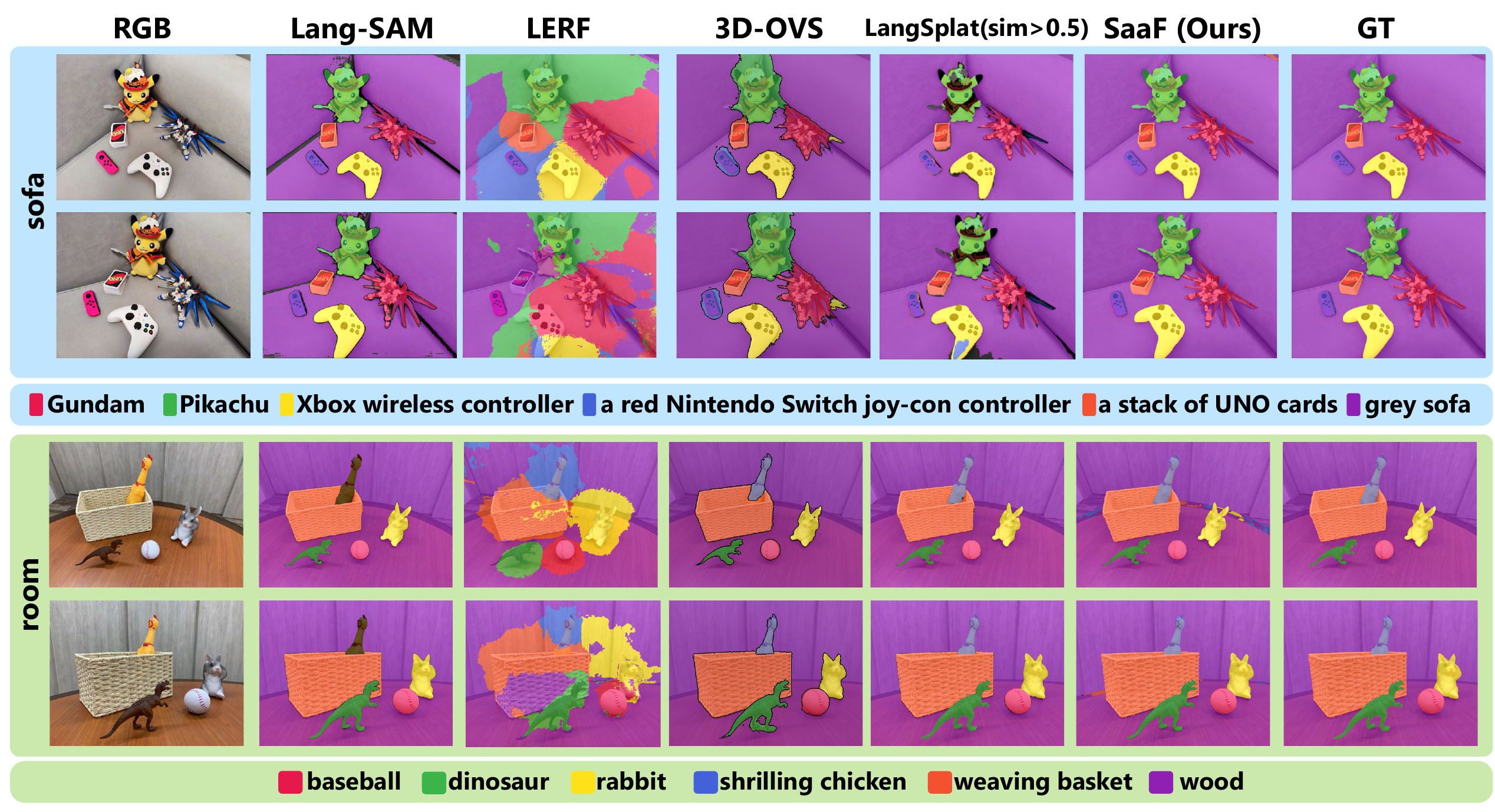}
    \caption{Qualitative results in 3D-OVS dataset. Segmentation result in the \textit{sofa} and \textit{room} scenes are visualized.}
    \label{fig:qualitative_results}
\end{figure*}

\Cref{fig:qualitative_results} presents qualitative results, and \cref{table:3d_ovs_results} shows IoU scores for the 3D-OVS dataset.
The proposed SaaF achieved the highest IoU accuracy in all scenes.
LERF exhibited unstable segmentation due to significant variation in features from different viewpoints.
LangSplat struggled in scenes like the \textit{sofa} scene, where the language features for the two similar controllers were close to each other, causing misrecognition.
SaaF, on the other hand, demonstrated robust recognition, even for objects sharing the same supercategory.

To test performance in more complex environments, we conducted additional experiments on the \textit{Figurines} scene from the LERF dataset, which features more objects and a more complex background.
\Cref{tab:lerf_dataset} shows the IoU evaluation results for this scene.
We confirmed that our method achieves higher accuracy compared to existing approaches in experiments using the LERF dataset.
Regarding ambiguity evaluation, we discuss recognition results for queries containing ambiguity, such as ducks with different accessories or different colored apples.

\begin{table*}[t]
    \centering
    \caption{Experimental results of 2D semantic segmentation on the 3D-OVS dataset. We resized the input images to $1,440\times 1,080$.}
    \begin{tabular}{l|ccc|ccc|c}
    \toprule
        & \multicolumn{3}{c|}{\textbf{w/o same supercategory objects}} &  \multicolumn{3}{c|}{\textbf{w/ same supercategory objects}} & ~ \\ \midrule
        Method & \textit{blue sofa} & \textit{lawn} & \textit{office desk} & \textit{bench} & \textit{room} & \textit{sofa} & \textit{overall} \\ \midrule
        Lang-SAM~\cite{langsam} & 55.0 & 89.4 & 57.8 & 89.3 & 71.3 & 85.1 & 74.6 \\
        LERF~\cite{lerf2023} & 46,6 & 73.7 & 47.3 & 53.2 & 46.6 & 27.0 & 49.6 \\ 
        3D-OVS~\cite{liu2023weakly} & 82.7 & 88.1 & 92.8 & 89.3 & 92.8 & 74.0 & 86.6 \\ 
        LangSplat~\cite{Qin_2024_CVPR} & 94.2 & 96.1 & 92.6 & 94.2 & 94.1 & 90.0 & 93.5 \\ \midrule
        \textbf{SaaF (Ours)} & \textbf{97.2} & \textbf{97.3} & \textbf{97.5} & \textbf{96.2} & \textbf{94.3} & \textbf{95.7} & \textbf{96.4} \\ 
        \bottomrule
    \end{tabular}
    \label{table:3d_ovs_results}
\end{table*}

\begin{table}[]
\caption{Experimental results on the LERF dataset.}
\begin{tabular}{l|rrrr}
\toprule
Scene & \multicolumn{1}{c}{Lang-SAM~\cite{langsam}} & \multicolumn{1}{c}{LERF~\cite{lerf2023}} & \multicolumn{1}{c}{LangSplat~\cite{Qin_2024_CVPR}} & \multicolumn{1}{c}{\textbf{SaaF}} \\
\midrule
\textit{Figurines}  &  19.5    & 38.6 &   44.7   &   \textbf{46.6}  \\
\bottomrule
\end{tabular}
\label{tab:lerf_dataset}
\end{table}


\subsection{Ambiguity Scores}
\begin{figure}
    \centering
    \includegraphics[width=1.0\linewidth]{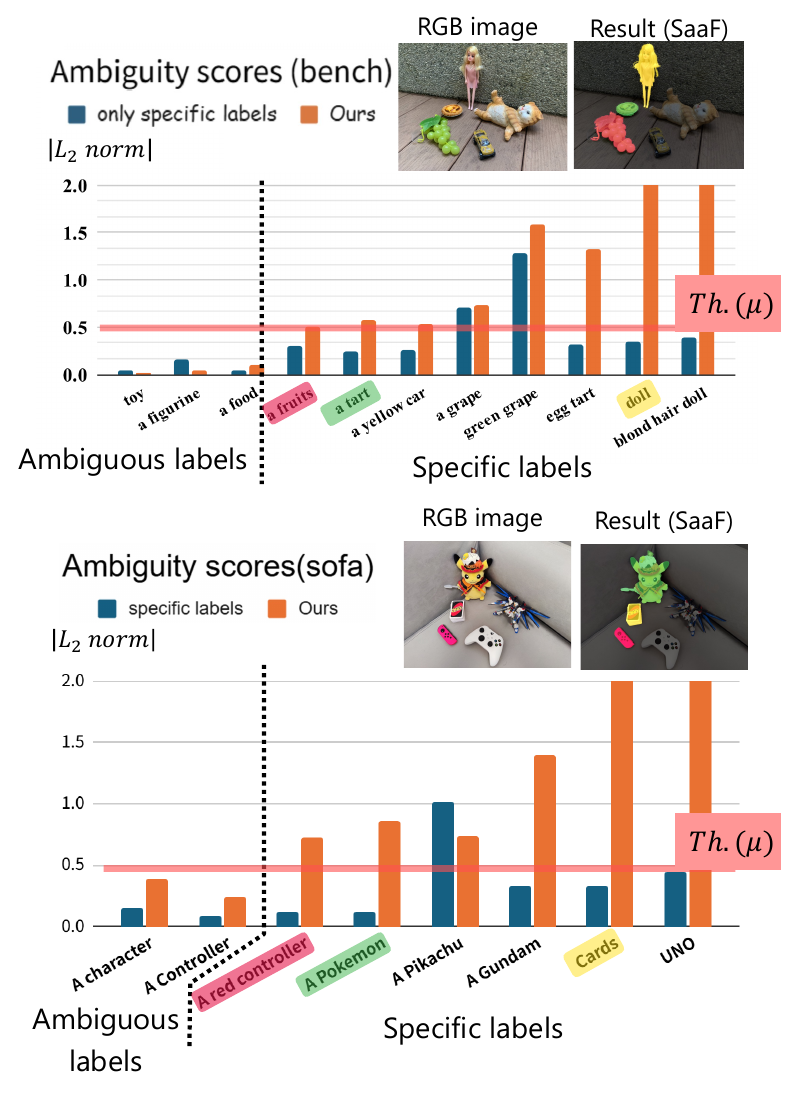}
    \caption{Comparison of L2 norm and segmentation results in \textit{bench} and \textit{sofa} scenes. The color applied to the query corresponds to the recognition result in the upper-right result.}
    \label{fig:ambiguity_scores_ovs}
\end{figure}

We further evaluated whether the proposed autoencoder can detect ambiguous queries.
We used the \textit{Figurines} scene from the LERF dataset, and the \textit{bench} and \textit{room} scenes from 3D-OVS, as they contain multiple objects from the same supercategory.
We define ambiguous queries as those with multiple valid targets, and unambiguous queries as those that point to a specific object.
We examined where these queries are mapped in the feature space.
\Cref{fig:ambiguity_scores_ovs} illustrates the results from the 3D-OVS dataset.
The blue bars in \cref{fig:ambiguity_scores_ovs} show the results of training without including ambiguous text labels, and the orange bars show the results of training with ambiguous text labels.
SaaF successfully separated ambiguous and unambiguous queries for the LERF dataset when using a threshold of 0.5.

In the \textit{bench} scene, the query \texttt{food} corresponds to both \texttt{tart} and \texttt{green grape}, and receives a score of 0.10 (below the threshold), indicating ambiguity.
Meanwhile, the query \texttt{fruit} scores 0.53, since only \texttt{green grape} is applicable in this scene.
We also confirmed that the more specific the query is, the higher the score becomes, 0.74 and 1.59, respectively.

When the text labels are not complex, the score for the queries \texttt{tart} and \texttt{fruit} is as low as 0.25, and the difference from the ambiguous query \texttt{figurine} is also tiny.

\begin{figure}
    \centering
    \includegraphics[width=1.0\linewidth]{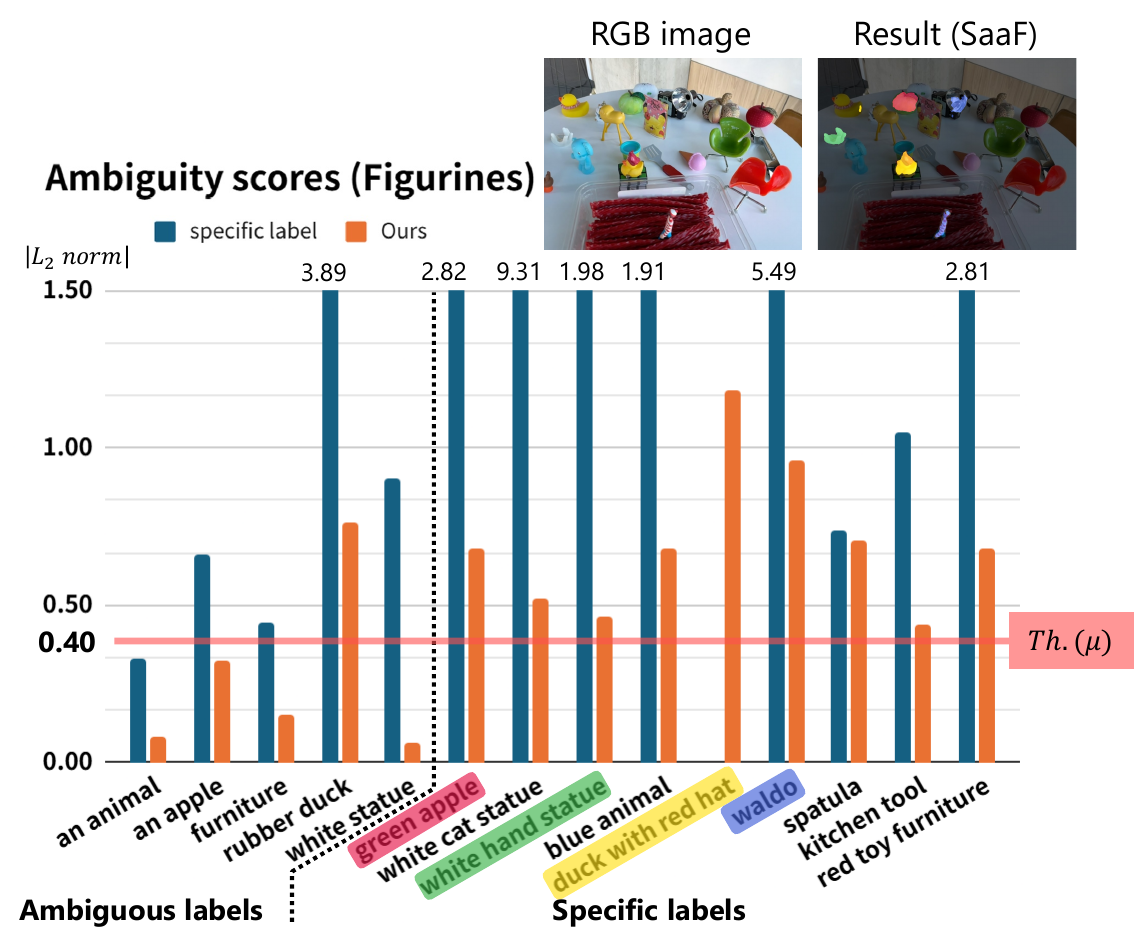}
    \caption{Comparison of L2 norm in \textit{Figurines}.}
    \label{fig:ambiguity_scores_lerf}
\end{figure}

\Cref{fig:ambiguity_scores_lerf} shows results of the \textit{Figurines} scene from the LERF dataset.
In this scene, we used labels of the objects arranged on the table.
For queries like \texttt{animal}, which matches multiple objects, the score was low at 0.08.
For queries like \texttt{an apple}, which contain object names, the score was 0.32, reflecting ambiguity due to the presence of both green and red apples.
More specific queries like \texttt{green apple} and \texttt{red apple} scored 0.68 and 0.60, respectively, clearly higher than a more general query of \texttt{an apple}.
The score of the supercategory \texttt{kitchen tool} is higher than the other supercategories (e.g., \texttt{animal}, \texttt{furniture}, \texttt{white statue}), 0.44, because the only kitchen tool in \textit{Figurine} is spatula. 
The score for \texttt{spatula} is 0.71, indicating that SaaF can correctly determine the ambiguity of the search query.

\subsection{Computational cost}
Object retrieval in LangSplat requires decoding features from all rendered pixels, taking an average of 9.6 [msec] in 3D-OVS dataset.
In contrast, SaaF only requires compressing query features, enabling search in 6.3 [msec], approximately 1.5 times faster than LangSplat.
The computational cost of object retrieval and ambiguity verification in SaaF is lower compared to LangSplat, making it a practical approach.

\subsection{Ablation Study}
We verify the effectiveness of each component of SaaF proposed in this paper: \textbf{best frame selection} and \textbf{multiple ambiguity labels} for training.
The specific text labels were generated by removing only the system prompt for generating ambiguous labels described in \cref{sec:label_generation}, and adding a prompt to generate approximately 10 distinct text labels per object instance.

\Cref{tab:component_ablation} shows the mIoU scores on the 3D-OVS dataset under different component settings. 
Without using any text labels, best frame selection alone improved segmentation accuracy by 9.0 points, indicating that the best frame selection can produce stable feature embedding in the 3D language field.
However, the mIoU score without text labels remained lower than existing methods.

When training with specific text labels describing detailed object features, the mIoU improves significantly by 37.5 points to 87.9\%.
This improvement suggests a strong association between image and text features in the trained feature space.

When training with multiple ambiguous text labels, the mIoU accuracy slightly decreased 1.5 points to 86.4\%.
In contrast, combining ambiguous text labels with best frame selection led to a significant improvement of 8.5 points, achieving an mIoU of 96.4\%. 

These results suggest that some frames contain ambiguity, and that using only the best frames helps reduce feature variation.
We confirmed that the highest accuracy was achieved when all components of SaaF were used together.

\begin{table}[]
\centering
\caption{Ablation results of 3D-OVS datasets.}


{\small
\begin{tabular}{cc|r}
\toprule
\multicolumn{2}{c|}{\textbf{Components}} & \multicolumn{1}{c}{\textbf{mIoU Accuracy (\%)}} \\ \midrule
Best Frame & Text-Label & \multicolumn{1}{c}{3D-OVS} \\
\midrule
~ & w/o                 & 41.4 \\
\checkmark & w/o      & 50.4 \\
\checkmark & Specific & 87.9 \\
~ & Multiple            & 86.4 \\
\textbf{\checkmark} & \textbf{Multiple} & \textbf{96.4} \\
\midrule
\end{tabular}
}

\label{tab:component_ablation}
\end{table}

\subsection{Limitation}
In this study, we used SAM2 to segment and track mask regions, and retrieval accuracy is highly dependent on the tracking accuracy.
SaaF learns a feature space in which CLIP features that cover multiple instances are detected as ambiguous.
As a result, if different instance IDs are assigned to the same object, all of its features may cluster near the origin.

In addition, SaaF does not yet handle the object hierarchies (e.g., whole, part, sub-part), which have been addressed in some related work~\cite{saga2023,Qin_2024_CVPR}.
Future works includes improving tracking performance by utilizing known camera parameters and enhancing the granularity of object retrieval to support hierarchical object understanding.

Another practical limitation is that the threshold $\mu$ must be determined manually based on the environment.
Real-world applications will require automatic threshold determination.

\section{Conclusion}

In this paper, we proposed SaaF, a 3D language field embedded with language features that enable scene-specific object retrieval and ambiguous query detection. 
For scene-specific object retrieval, we proposed a metric learning-based autoencoder to separate the features of each object. 
The feature space can also determine the ambiguity of a search query by ambiguity-aware training using generated text labels with multi-level ambiguity for each object instance.
Experimental results showed that the proposed method provides the highest accuracy in object retrieval compared to 3D language fields based on NeRF and Gaussian Splatting. 
We also compared the size of compressed features obtained from text queries and showed that it is possible to determine ambiguity.
These experimental results indicate SaaF can compute ambiguity and retrieve an object simultaneously in scenarios containing ambiguity, such as \cref{fig:target_scnario}.
By applying the proposed method to robot tasks with user interaction, we believe it is possible to realize more practical service robots that account for ambiguity.
In future work, we will apply SaaF to robot tasks in real-world environments with user interaction.


\bibliographystyle{ieeetr}
\bibliography{ref.bib}

@article{ravi2024sam2,
  title={SAM 2: Segment Anything in Images and Videos},
  author={Ravi, Nikhila and Gabeur, Valentin and Hu, Yuan-Ting and Hu, Ronghang and Ryali, Chaitanya and Ma, Tengyu and Khedr, Haitham and R{\"a}dle, Roman and Rolland, Chloe and Gustafson, Laura and Mintun, Eric and Pan, Junting and Alwala, Kalyan Vasudev and Carion, Nicolas and Wu, Chao-Yuan and Girshick, Ross and Doll{\'a}r, Piotr and Feichtenhofer, Christoph},
  journal={arXiv preprint arXiv:2408.00714},
  year={2024}
}

@inproceedings{wu2024opengaussian,
title={{OpenGaussian}: Towards point-level {3D} {Gaussian}-based open vocabulary understanding},
author={Yanmin Wu and Jiarui Meng and Haijie LI and Chenming Wu and Yahao Shi and Xinhua Cheng and Chen Zhao and Haocheng Feng and Errui Ding and Jingdong Wang and Jian Zhang},
booktitle={Proceedings of the 38th Annual Conference on Neural Information Processing Systems},
year={2024},
}

@inproceedings{shi2024language,
  title={Language embedded {3D} gaussians for open-vocabulary scene understanding},
  author={Shi, Jin-Chuan and Wang, Miao and Duan, Hao-Bin and Guan, Shao-Hua},
  booktitle={Proceedings of the 2024 IEEE/CVF Conference on Computer Vision and Pattern Recognition},
  pages={5333--5343},
  year={2024}
}

@inproceedings{lerf2023,
 author = {Kerr, Justin and Kim, Chung Min and Goldberg, Ken and Kanazawa, Angjoo and Tancik, Matthew},
 title = {{LERF}: Language Embedded Radiance Fields},
 booktitle = {Proceedings of the 19th IEEE/CVF International Conference on Computer Vision},
pages= {19729--19739},
 year = {2023},
}

@InProceedings{Qin_2024_CVPR,
    author    = {Qin, Minghan and Li, Wanhua and Zhou, Jiawei and Wang, Haoqian and Pfister, Hanspeter},
    title     = {{LangSplat}: {3D} Language Gaussian Splatting},
    booktitle = {Proceedings of the 2024 IEEE/CVF Conference on Computer Vision and Pattern Recognition},
    month     = {June},
    year      = {2024},
    pages     = {20051--20060}
}

@InProceedings{clip_pmlr-v139-radford21a,
  title = 	 {Learning Transferable Visual Models From Natural Language Supervision},
  author =       {Radford, Alec and Kim, Jong Wook and Hallacy, Chris and Ramesh, Aditya and Goh, Gabriel and Agarwal, Sandhini and Sastry, Girish and Askell, Amanda and Mishkin, Pamela and Clark, Jack and Krueger, Gretchen and Sutskever, Ilya},
  booktitle = 	 {Proceedings of the 38th International Conference on Machine Learning},
  pages = 	 {8748--8763},
  year = 	 {2021},
  editor = 	 {Meila, Marina and Zhang, Tong},
  volume = 	 {139},
}

@inproceedings{ren2023robots,
title={Robots That Ask For Help: Uncertainty alignment for large language model planners},
author={Allen Z. Ren and Anushri Dixit and Alexandra Bodrova and Sumeet Singh and Stephen Tu and Noah Brown and Peng Xu and Leila Takayama and Fei Xia and Jake Varley and Zhenjia Xu and Dorsa Sadigh and Andy Zeng and Anirudha Majumdar},
booktitle={Proceedings of the 7th Annual Conference on Robot Learning},
year={2023},
}

@inproceedings{ye2024gaussian,
  title={Gaussian grouping: Segment and edit anything in {3D} scenes},
  author={Ye, Mingqiao and Danelljan, Martin and Yu, Fisher and Ke, Lei},
  booktitle={Computer Vision --- ECCV 2024},
  pages={162--179},
  year={2024},
}

@article{saga2023,
  publtype={informal},
  author={Cen, Jiazhong and Fang, Jiemin and Yang, Chen and Xie, Lingxi and Zhang, Xiaopeng and Shen, Wei and Tian, Qi},
  title={Segment Any {3D} {Gaussians}},
  year={2023},
  cdate={1672531200000},
  journal={CoRR},
  volume={abs/2312.00860},
}

@inproceedings{choi2024click,
  title={Click-Gaussian: Interactive segmentation to any {3D} {Gaussians}},
  author={Choi, Seokhun and Song, Hyeonseop and Kim, Jaechul and Kim, Taehyeong and Do, Hoseok},
  booktitle={Computer Vision --- ECCV 2024},
  pages={289--305},
  year={2025},
  organization={Springer}
}

@Article{kerbl3Dgaussians,
      author       = {Kerbl, Bernhard and Kopanas, Georgios and Leimk{\"u}hler, Thomas and Drettakis, George},
      title        = {{3D} Gaussian Splatting for Real-Time Radiance Field Rendering},
      journal      = {ACM Transactions on Graphics},
      number       = {4},
      volume       = {42},
      month        = {July},
      year         = {2023},
}

@article{nerf,
author = {Mildenhall, Ben and Srinivasan, Pratul P. and Tancik, Matthew and Barron, Jonathan T. and Ramamoorthi, Ravi and Ng, Ren},
title = {{NeRF}: Representing scenes as neural radiance fields for view synthesis},
year = {2021},
issue_date = {January 2022},
address = {New York, NY, USA},
volume = {65},
number = {1},
issn = {0001-0782},
journal = {Communications of the ACM},
pages = {99--106},
numpages = {8}
}

@inproceedings{li2022languagedriven,
title={Language-driven Semantic Segmentation},
author={Li, Boyi and Weinberger, Q. Kilian and Belongie, Serge and Koltun, Vladlen and Ranftl, Rene},
booktitle={Proceedings of the 2022 International Conference on Learning Representations},
year={2022},
}

@inproceedings{liu2024grounding,
  title={Grounding {DINO}: Marrying {DINO} with grounded pre-training for open-set object detection},
  author={Liu, Shilong and Zeng, Zhaoyang and Ren, Tianhe and Li, Feng and Zhang, Hao and Yang, Jie and Jiang, Qing and Li, Chunyuan and Yang, Jianwei and Su, Hang and others},
  booktitle={Computer Vision --- ECCV 2024},
  pages={38--55},
  year={2025},
  organization={Springer}
}

@inproceedings{kirillov2023segment,
  title={Segment anything},
  author={Kirillov, Alexander and Mintun, Eric and Ravi, Nikhila and Mao, Hanzi and Rolland, Chloe and Gustafson, Laura and Xiao, Tete and Whitehead, Spencer and Berg, Alexander C and Lo, Wan-Yen and others},
  booktitle={Proceedings of the 19th IEEE/CVF International Conference on Computer Vision},
  pages={4015--4026},
  year={2023}
}

@article{liu2023weakly,
  title={Weakly supervised {3D} open-vocabulary segmentation},
  author={Liu, Kunhao and Zhan, Fangneng and Zhang, Jiahui and Xu, Muyu and Yu, Yingchen and El Saddik, Abdulmotaleb and Theobalt, Christian and Xing, Eric and Lu, Shijian},
  journal={Advances in Neural Information Processing Systems},
  volume={36},
  pages={53433--53456},
  year={2023}
}

@article{dai2023instructblip,
  title={{InstructBLIP}: Towards General-purpose Vision-Language Models with Instruction Tuning},
  author={Dai, Wenliang and Li, Junnan and LI, DONGXU and Tiong, Anthony and Zhao, Junqi and Wang, Weisheng and Li, Boyang and Fung, Pascale N and Hoi, Steven},
  journal={Advances in Neural Information Processing Systems},
  volume={36},
  pages={49250--49267},
  year={2023}
}

@article{liu2023visual,
  title={Visual instruction tuning},
  author={Liu, Haotian and Li, Chunyuan and Wu, Qingyang and Lee, Yong Jae},
  journal={Advances in neural information processing systems},
  volume={36},
  pages={34892--34916},
  year={2023}
}

@article{ahn2022can,
  title={Do as {I} can, not as {I} say: Grounding language in robotic affordances},
  author={Ahn, Michael and Brohan, Anthony and Brown, Noah and Chebotar, Yevgen and Cortes, Omar and David, Byron and Finn, Chelsea and Fu, Chuyuan and Gopalakrishnan, Keerthana and Hausman, Karol and others},
  journal={arXiv preprint arXiv:2204.01691},
  year={2022}
}

@inproceedings{yano2024unified,
  title={Unified understanding of environment, task, and human for human-robot interaction in real-world environments},
  author={Yano, Yuga and Mizutani, Akinobu and Fukuda, Yukiya and Kanaoka, Daiju and Ono, Tomohiro and Tamukoh, Hakaru},
  booktitle={Proceedings of the 33rd IEEE International Conference on Robot and Human Interactive Communication},
  pages={224--230},
  year={2024},
}

@inproceedings{zitkovich2023rt,
  title={{RT-2}: Vision-language-action models transfer {Web} knowledge to robotic control},
  author={Zitkovich, Brianna and Yu, Tianhe and Xu, Sichun and Xu, Peng and Xiao, Ted and Xia, Fei and Wu, Jialin and Wohlhart, Paul and Welker, Stefan and Wahid, Ayzaan and others},
  booktitle={Proceedings of the 7th Conference on Robot Learning},
  pages={2165--2183},
  year={2023},
}

@misc{langsam,
  author = {Luca Medeiros},
  title = {Language-segment anything},
  note = {\url{https://github.com/luca-medeiros/lang-segment-anything} (accessed 8. Apr. 2025)},
  year = {2024}
}

@inproceedings{caron2021emergingDINO,
  title={Emerging properties in self-supervised vision transformers},
  author={Caron, Mathilde and Touvron, Hugo and Misra, Ishan and J{\'e}gou, Herv{\'e} and Mairal, Julien and Bojanowski, Piotr and Joulin, Armand},
  booktitle={Proceedings of the 2021 IEEE/CVF International Conference on Computer Vision},
  pages={9650--9660},
  year={2021}
}

@inproceedings{azuma_2022_CVPR,
  title={{ScanQA}: {3D} Question Answering for Spatial Scene Understanding},
  author={Azuma, Daichi and Miyanishi, Taiki and Kurita, Shuhei and Kawanabe, Motoaki},
  booktitle={Proceedings of the 2022 IEEE/CVF Conference on Computer Vision and Pattern Recognition},
  year={2022},
  pages={19129--19139}
}

@INPROCEEDINGS{sim_vqa_2022,
  author={Cascante-Bonilla, Paola and Wu, Hui and Wang, Letao and Feris, Rogerio and Ordonez, Vicente},
  booktitle={Proceedings of the 2022 IEEE/CVF Conference on Computer Vision and Pattern Recognition}, 
  title={{Sim VQA}: Exploring Simulated Environments for Visual Question Answering}, 
  year={2022},
  pages={5056--5066},
}

@misc{gpt4v,
    author={OpenAI},
    title = {{GPT-4v(ision)} system card},
    note = {\url{https://cdn.openai.com/papers/GPTV_System_Card.pdf}},
    year = {2023}
}

\end{document}